\documentclass[runningheads]{llncs}
\usepackage{graphicx}
\usepackage{comment}
\usepackage{amsmath,amssymb} 
\usepackage{epsfig}
\usepackage{graphicx}
\usepackage{amsmath}
\usepackage{amssymb}
\usepackage{mwe}
\usepackage{color}
\usepackage{xpatch}
\usepackage{subcaption}
\usepackage{soul}
\usepackage{url}
\usepackage[noadjust]{cite}
\usepackage{wrapfig,lipsum,booktabs}
\newcommand{\etal}{\textit{et al.}}

\begin{document}
	\pagestyle{headings}
	\mainmatter
	\def\ECCVSubNumber{2920}  
	
	\title{Collaborative Training between Region Proposal Localization and Classification for Domain Adaptive Object Detection} 

	\titlerunning{Collaborative Training for Domain Adaptive Object Detection}
	%
	\author{Ganlong Zhao \and
	Guanbin Li\thanks{Corresponding author is Guanbin Li.} \and
	Ruijia Xu \and Liang Lin}
	\authorrunning{G. Zhao et al.}
	%
	\institute{School of Data and Computer Science, Sun Yat-sen University, Guangzhou, China
	\email{zhaoglong@mail2.sysu.edu.cn, liguanbin@mail.sysu.edu.cn, xurj3@mail2.sysu.edu.cn, linliang@ieee.org}}
	\maketitle
	
	\begin{abstract}
		Object detectors are usually trained with large amount of labeled data, which is expensive and labor-intensive. Pre-trained detectors applied to unlabeled dataset always suffer from the difference of dataset distribution, also called domain shift. Domain adaptation for object detection tries to adapt the detector from labeled datasets to unlabeled ones for better performance. In this paper, we are the first to reveal that the region proposal network~(RPN) and region proposal classifier~(RPC) in the endemic two-stage detectors~(e.g., Faster RCNN) demonstrate significantly different transferability when facing large domain gap. The region classifier shows preferable performance but is limited without RPN's high-quality proposals while simple alignment in the backbone network is not effective enough for RPN adaptation. We delve into the consistency and the difference of RPN and RPC, treat them individually and leverage high-confidence output of one as mutual guidance to train the other. Moreover, the samples with low-confidence are used for discrepancy calculation between RPN and RPC and minimax optimization. Extensive experimental results on various scenarios have demonstrated the effectiveness of our proposed method in both domain-adaptive region proposal generation and object detection. Code is available at \url{https://github.com/GanlongZhao/CST_DA_detection}.


		\keywords{Domain adaptation, Object detection, Transfer learning.}
	\end{abstract}

	\section{Introduction}
	Benefiting from massively well-labeled data, deep convolutional neural networks have recently shown unparalleled advantages in various visual tasks, e.g., image recognition and object detection. Unfortunately, such data is usually prohibitive in many real-world scenarios. The problem becomes extremely serious for object detection, since it requires more precise object-level annotations. 
	A common solution for this problem is to transfer the pretrained model from label-rich domain(i.e. source domain) to the other(i.e. target domain), but this often suffers from performance degradation due to domain gap.

	\begin{figure}[t]
		\begin{center}
			\includegraphics[width=0.5\linewidth]{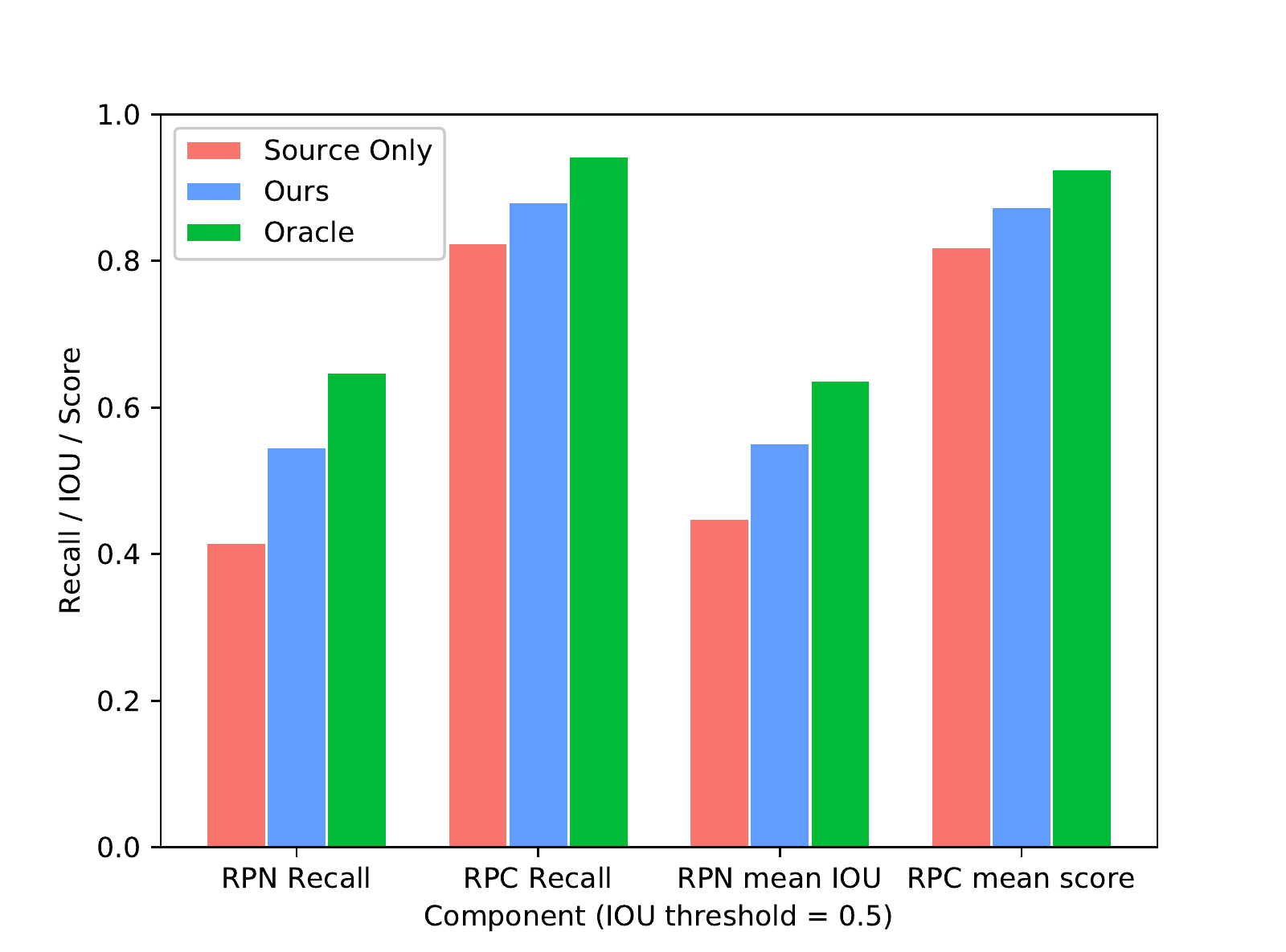}
		\end{center}
		\caption{Comparison of domain adaptation performance between ``Source Only'' model, ``Ours'' and the ``Oracle'' model w.r.t ``Recall'', ``Average IOU'' and ``Mean Score''. 
			Threshold of IOU is set to 0.5 for ``Recall'' calculation. 
			Experiment is conducted on adapting models trained on Sim10k~\cite{johnson2016driving} to Cityscapes~\cite{cordts2016cityscapes}.}
		\label{rpn_cls_compare}
	\end{figure}
	Various kinds of methods have been proposed to overcome the domain gap. Most of them are based on adversarial training, and can be separated into two categories: feature-level and pixel-level domain adaptation. Feature-level domain adaptation tries to align the feature distributions from the two domains by adversarial training, while pixel-level domain adaptation uses GANs~\cite{goodfellow2014generative} to generate target-like images from source domain with labels unchanged. 
	
	There have been some research works focusing on domain adaptive object detection with feature-level or pixel-level adaptation techniques. Chen \etal~\cite{chen2018domain} attempt to reduce domain discrepancy in both image level and instance level. Saito \etal~\cite{saito2018strong} leverage weak global feature alignment and strong local alignment to mitigate the performance degradation caused by distinct scene layouts and different combinations of objects. Zhu \etal~\cite{zhu2019adapting} propose to bridge the domain gap through effective region mining and region-based domain alignment. 
	Noted that most of the previous research works on domain adaptive object detection focus on bridging the whole-image representations and thus perform alignment in the backbone branch of a detector. Though Zhu \etal~\cite{zhu2019adapting} propose a selective adaptation framework based on region patches generated by the RPN branch, the loss gradient of the region classifier is just propagated to the backbone without adapting the RPN itself. Saito \etal~\cite{saito2018strong} conduct feature alignment only on the lower and final layer of the backbone. However, different from neural network based classification models, most of endemic two-stage object detectors are far more complex. It is far from sufficient to align and adapt the global features of the backbone network, which ignores the transferability of the RPN module.


	RPN transferability is paramount for the adaptation of two-stage detectors, while adapting on the entire image with backbone-only alignment is not an effective solution. A two-stage object detector can be separated into three modules: backbone network, RPN and region proposal classifier(abbr.~``RPC''). With large domain gap and complicated scene, we empirically discover that RPN and RPC show different transferability, i.e., RPC usually performs better than RPN. We adopt the ``Source Only'' domain adaptation model to investigate the transferability difference of RPN and RPC. Specifically, we directly apply the model trained on Sim10k~\cite{johnson2016driving} to test the performance on Cityscapes~\cite{cordts2016cityscapes}, take 0.5 as the IOU threshold and compute the recall of RPN and RPC respectively~\footnote{Noted that the recall computation of RPC here refers to the proportion of detected ROIs (not GT objects) having correct label prediction.}. As shown in Fig.~\ref{rpn_cls_compare}, 
	it is obvious that RPC performs better than RPN before and after adaptation, and more importantly, has much less degradation between oracle(green bar) and the source-only model(red bar), which implies that RPN suffers much severer than RPC from domain gap. However, the performance of RPC is also limited if RPN fails to provide high-quality region proposals. RPN has therefore become the bottleneck. Noted that RPC here is not doomed to be better than RPN, because it considers the classification recall of the detected region proposals~(even if the RPN detection is accurate, the accuracy of the proposal classification is still uncertain).


	On the other hand, it is noteworthy that, in some kinds of two-stage object detectors~(e.g., Faster RCNN), there is no gradient flow between RPN and RPC. A natural idea is to take them as two individual branches from backbone. If we consider RPN as a kind of foreground/background classifier, it can be regarded as a coarse and fast RPC classifying each anchor across the feature map. Similarly, if we sum up the output of RPC stream to background and foreground scores, it performs just like a fine-grained and selective RPN. Based on the above discussion, we propose a novel domain adaptation method on Faster RCNN using collaborative training between RPN and RPC. It can also be easily generalized to other two-stage detectors. Specifically, we first apply collaborative self-training between RPN and RPC, which leverages the high-confident output of one to train the other. Besides, we introduce focal loss~\cite{lin2017focal} in our method to impose more weight on ROIs of high-confidence and improve stability by removing the threshold selection. Second,  ROIs of low-confidence that are ignored in the first part are used to calculate the foreground/background discrepancy between RPN and RPC. 
	To improve the detector's transferability, the backbone network is trained to minimize the discrepancy while RPN and RPC try to maximize it. We verify its effectiveness under different adaptation scenarios.
	
	To sum up, this work has the following contributions: (1) We are the first to reveal the significance of exploring the transferability of RPN module for domain-adaptive object detection. Simple alignment in backbone can not guarantee that the RPC receives high quality proposals. (2) From the perspective of treating RPN and RPC independently, we derive a collaborative self-training method that can propagate the loss gradient through the whole network and mutually enhance each other. (3) To the best of our knowledge, we are the first to adapt Maximum Classifier Discrepancy, MCD~\cite{saito2018maximum} to two-stage object detection framework for domain adaptation by focusing on ambiguous ROIs and show its effectiveness.

	\section{Related Works}
	
	\paragraph{\textbf{Object Detection}}
	
	Object detection has been around for a long time and is now an important research topic in computer vision. The development of convolutional neural networks has greatly advanced the performance of object detection. CNN-based detectors can be mainly divided into two categories: one-stage detectors and two-stage detectors. Although one-stage detectors such as YOLO~\cite{redmon2016you} and SSD~\cite{liu2016ssd} have notably higher efficiency and have become popular paradigms, two-stage detectors like Fast RCNN~\cite{girshick2015fast}, Faster RCNN~\cite{ren2015faster} and Mask RCNN~\cite{he2017mask} are still widely adopted for their much higher performance. Faster RCNN~\cite{ren2015faster} is a classical two-stage object detector and is commonly used as a baseline for domain adaptation. 
	However, object detectors suffer from domain gap when being applied to an unseen domain. Generally the backbone network pre-trained with ImageNet~\cite{deng2009imagenet} is fine-tuned on large amount of object-level labeled data for detection together with RPN and RPC. Unfortunately, such annotated data is usually prohibitive in target domains. 
	
	\paragraph{\textbf{Domain Adaptation}}	
	Domain adaptation aims to utilize the labeled source domain data and unlabeled target domain data to boost performance on the latter. Domain adaptation on classification has been widely studied with technical paradigms like subspace alignment~\cite{fernando2013unsupervised}, asymmetric metric learning~\cite{kulis2011you} and covariance matrix alignment~\cite{sun2016return}. A typical approach for domain adaptation is to reduce domain gap by making features or images from the two domains indistinguishable. Some methods try to minimize the distance between features of the two domains by resorting to MMD~\cite{borgwardt2006integrating} or $\mathcal{H}$\text{$\Delta$}$\mathcal{H}$~\cite{saito2018maximum}, while some of the other works employ adversarial training with gradient reverse layer~\cite{ganin2016domain} or use generative adversarial networks~\cite{bousmalis2017unsupervised, yoo2016pixel}. Besides, entropy minimization has also been applied to domain adaptation for classification~\cite{long2016unsupervised} and segmentation~\cite{vu2019advent}.  Domain-adaptive object detection has a completely different framework from image classification and semantic segmentation. It includes both object proposal detection and region-level classification. Therefore, when designing a domain-adaptive detection algorithm, it is far from sufficient to simply consider the alignment of the backbone network features, and it is necessary to consider the transferability of the algorithm in both RPN and RPC.
	
	\paragraph{\textbf{Domain Adaptation for Object Detection}}
	In the past few years there has been some research in domain adaptation for object detection~\cite{hoffman2017cycada, kim2019diversify, roychowdhury2019automatic, tzeng2018splat, wang2019few}. Raj~\etal~\cite{raj2015subspace} first proposed a domain adaptation method on RCNN with subspace alignment.
	Chen~\etal~\cite{chen2018domain} used a global-level and an instance-level alignment method respectively for the global and regional features. Inoue~\etal~\cite{inoue2018cross} proposed a weakly-supervised framework with pseudo label. Saito~\etal~\cite{saito2018strong} pointed out that global alignment on complicated scenes and layouts might lead to negative transfer, and they proposed to employ a weak global alignment in the final layer of the backbone network, which puts more emphasis on images that are globally similar, and a strong local alignment in lower layer of the backbone. Zhu~\etal~\cite{zhu2019adapting} utilized RPN proposals to mine the discriminative regions of fixed size, which are pertinent to object detection, and focused on the alignment of those regions. These methods mainly focus on the alignment of the backbone stream regardless of the RPN transferability. 

	\section{Method}
	\label{section_method}
	
	\subsection{Framework overview}
	
	Given one labeled dataset from the source domain and an unlabeled one from the target domain, our task is to train a detector to obtain the best performance on the target dataset. For simplicity, both datasets share the same label space. Traditionally, feature-level domain adaptation methods try to extract the domain-invariant feature from both datasets, neglecting the adaptation of the main modules~(i.e., RPN and RPC) besides the backbone stream.	
	
\begin{figure}[t]
		\begin{center}
			\includegraphics[width=0.5\linewidth]{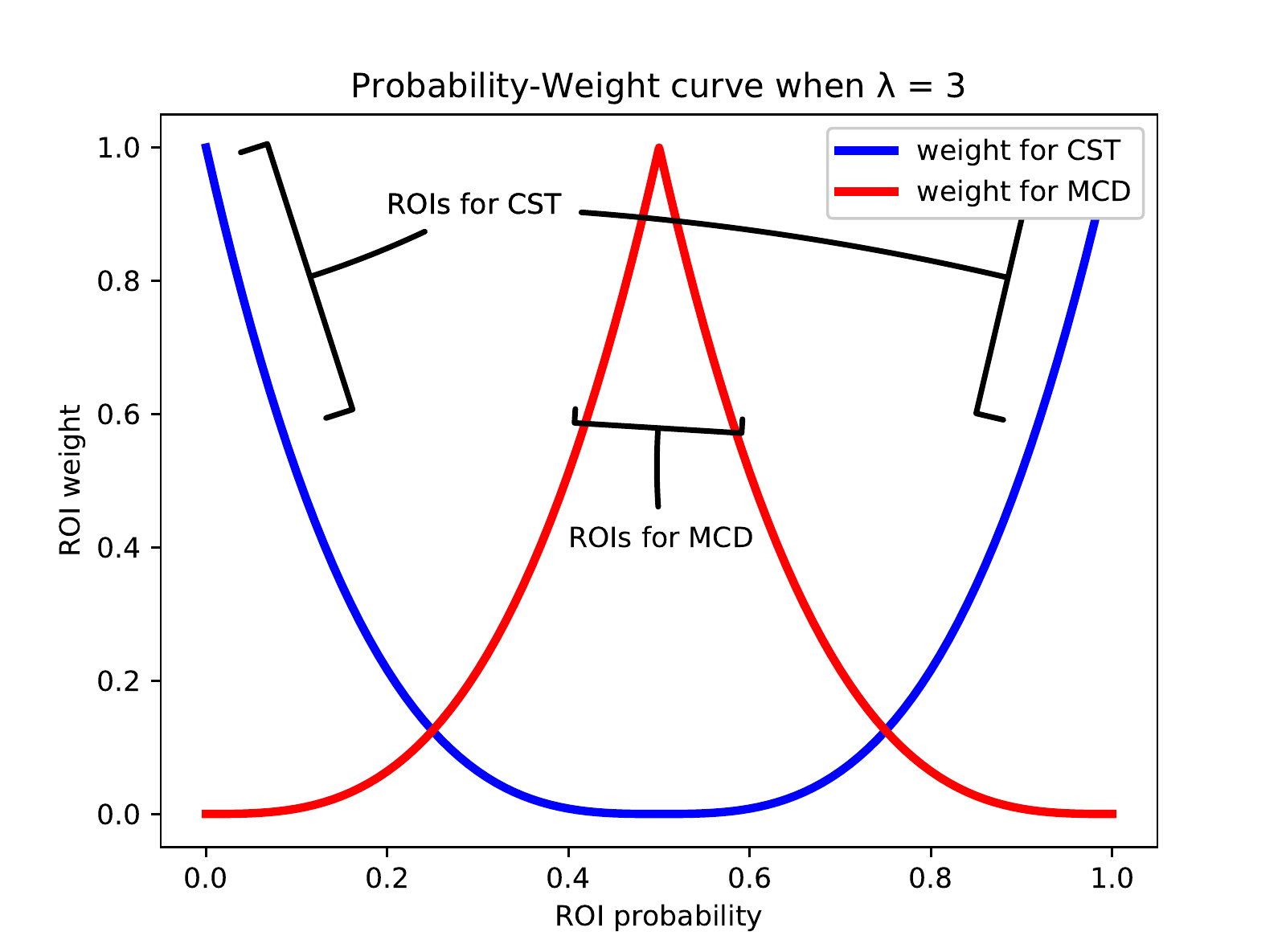}
		\end{center}
		
		\caption{The weight of ROI w.r.t its probability in loss calculation. 
			Blue curve is used for collaborative self-training, and the red curve is for MCD.}
		\label{weightcurve}
	\end{figure}

\begin{figure*}[t]
		\begin{center}
			\includegraphics[width=0.7\linewidth]{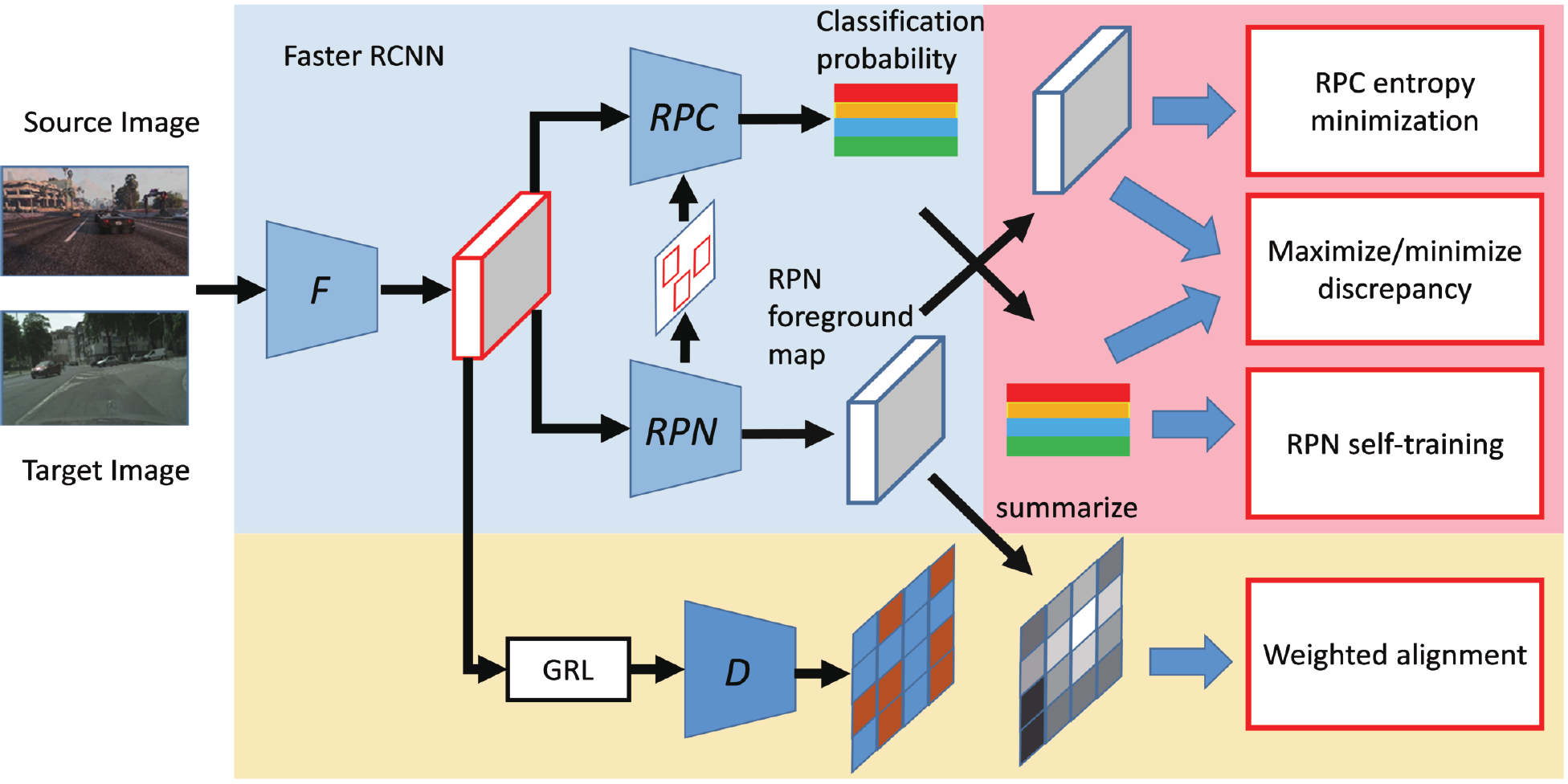}
		\end{center}
		\caption{Network architecture of our method. GRL stands for gradient reverse layer. From top to bottom: entropy minimization on RPC probability weighted by corresponding foreground score in RPN foreground map; discrepancy calculation on ambiguous ROIs shared in both RPN and RPC; RPN self-training using high-confident RPC detection results; domain discriminator loss weighted by summarized foreground probability in the same position.}
		\label{network}
	\end{figure*}

	The architecture of our model is illustrated in Fig.~\ref{network}. The blue region on the top includes the modules in Faster RCNN, in which RPN generates and sends ROIs to the head of RPC for ROI-pooling. The yellow part is a domain discriminator that tries to determine which domain the input features originate from. It takes the backbone feature as input and outputs a domain prediction map of the same size. Besides, RPN prediction is used to highlight the foreground anchors in discriminator loss calculation. The red part consists of the proposed collaborative self-training scheme and the discrepancy maximization/minimization between RPN and RPC. ROIs with high-confidence of RPN~(RPC) are used to train RPC~(RPN), while ambiguous ROIs are used for discrepancy optimization. We illustrate the mutual complementary relationship of the proposed collaborative training and MCD optimization in Fig.~\ref{weightcurve}. Among them, for collaborative training, the higher the confidence level of the ROI, the greater the weight will be given when calculating the loss function, while the opposite is true for MCD optimization. The lower the confidence level of ROI, the larger the sample weight will be. The two curves are implemented by tailor-designed polynomial functions in our experiment.

	
	\subsection{Collaborative self-training}
	
	Generally a prevailing two-stage detector can be separated into three parts: the backbone F, RPN and RPC. Backbone F plays the role of feature representation and extracts the feature of the whole image. Then RPN takes the feature as input, and predicts the foreground/background score of each anchor across the feature map. ROI pooling is applied to the anchors of high foreground probability for feature extraction and further sent to the RPC. Finally, RPC performs category prediction and regression of size and position of bounding boxes. 
	
	Although RPC performs regional classification based on the resulted proposals of the RPN module, it does not back propagate gradient to RPN during training. RPN filters out anchors with low foreground probability before feeding the rest to the RPC module. 
	If the proposal filtering operation is removed and the RPN module performs ROI pooling at each anchor, the RPC module can be considered equivalent to the RPN. Ideally, the outputs of RPN and RPC should also be consistent. Those anchors with high background score in RPC should have low RPN foreground probability. Similarly, anchors having high score with non-background classes should also have high RPN foreground probability.
	
	The core motivation of this paper is to improve the performance of object detection in the target domain by fully exploiting the domain-adaptive capability of the RPN module. Now we have accessed to the labeled image~\begin{math}x_s\end{math} and its annotation~\begin{math}y_s\end{math} from an annotated source dataset~\begin{math}\{X_s, Y_s\}\end{math}, as well as the unlabeled image~\begin{math}x_t\end{math} drawn from an unlabeled target dataset~\begin{math}\{X_t\}\end{math}. For labeled source image~\begin{math}x_s\end{math}, we introduce supervised training loss of Faster RCNN, which is calculated as follows~\cite{chen2018domain}:
	\begin{equation}
	L_{det} = L_{rpn}(x_s, y_s) + L_{cls}(x_s, y_s).
	\end{equation}
	
	As we do not have accessed to the annotations~\begin{math}\{y_t\}\end{math}, we mutually train the two modules of RPN and RPC by leveraging the output of one module to train the other. Given a target image~\begin{math}x_t\end{math}, feature~\begin{math}f_t\end{math} is first extracted by the feature extractor \begin{math}F\end{math}. Based on~\begin{math}f_t\end{math}, RPN predicts the score~(i.e., foreground and background probability)~\begin{math}s_{rpn}\end{math} for each ROI~\begin{math}r_t\end{math} while RPC outputs its class probability distribution~\begin{math}s_{cls}\end{math}, including the score of the background category and several foreground ones. For those ROIs with high-confident $s_{cls}$, we reuse them to update RPN, the loss of which is calculated as:
	\begin{equation}
	\label{L_rpn_t}
	L_{rpn_t} = f_w(s_{cls}) L_{rpn}(x_t, \hat{y_t}),
	\end{equation} where \begin{math}f_w\end{math} can be defined as any function that decreases when \begin{math}s_{cls}\end{math} is uncertain on the foreground/background classification. For simplicity, we define it as:
	\begin{equation}
	f_w(s_{cls}) = (| 1 - 2s_{cls}^{bg} |) ^\lambda,
	\end{equation} where \begin{math} s_{cls}^{bg}\end{math} is the background score in \begin{math}s_{cls}\end{math}. $\lambda$ controls the weight on samples of low-confidence. Besides, \begin{math}\hat{y}_t\end{math} in Eq~\ref{L_rpn_t} refers to the pseudo label which contains ROIs with high-confident \begin{math}s_{cls}\end{math}, including both foreground and background region proposals. It does not need to contain every object in the original image nor the specific class. In Faster RCNN, RPN is trained in a selective way. Assuming that the feature extractor forwards a feature map of spatial size 
	\begin{math}H\times W\end{math}, there will generally be \begin{math}H \times W \times 9\end{math} anchors and a prediction map of equal size. Only a small portion of anchors are referenced in \begin{math}L_{rpn}\end{math} calculation and missing labels do not hurt the performance of RPN without extra processing.
	
	
	On the other side, we can perform similar operations in the RPC module. Since RPN focuses on foreground/background classification and can not provide anchors with category-level pseudo labels which is necessary for RPC training, we adopt entropy minimization~\cite{long2016unsupervised} for RPC, and adaptively assign higher weight to the samples of high-confidence in the calculation of the loss function. Based on entropy minimization, RPC is trained to output high-confident \begin{math}s_{cls}\end{math} for ROIs with high-confident \begin{math}s_{rpn}\end{math}. Similar to the RPN module, we define:
	\begin{equation}
	L_{cls_t} = f_w(s_{rpn}) E(s_{cls}),
	\end{equation}
	\begin{equation}
	E(s_{cls}) = -\sum_{c \in C}{s_{cls}^c log({s_{cls}^c)}},
	\end{equation}
	\begin{equation}
	f_w(s_{rpn}) = |1 - 2s_{rpn}^{fg}|^\lambda,
	\end{equation}
	where \begin{math}C\end{math} includes the background and all foreground classes. \begin{math}s_{cls}^c\end{math} denotes the predicted probability of class \begin{math}c\end{math} in \begin{math}s_{cls}\end{math} while \begin{math}s_{rpn}^{fg}\end{math} refers to the output foreground probability of the RPN module.

	\subsection{Maximize discrepancy classifier on detectors}
\begin{figure}[t]
		\begin{center}
			\includegraphics[width=0.6\linewidth]{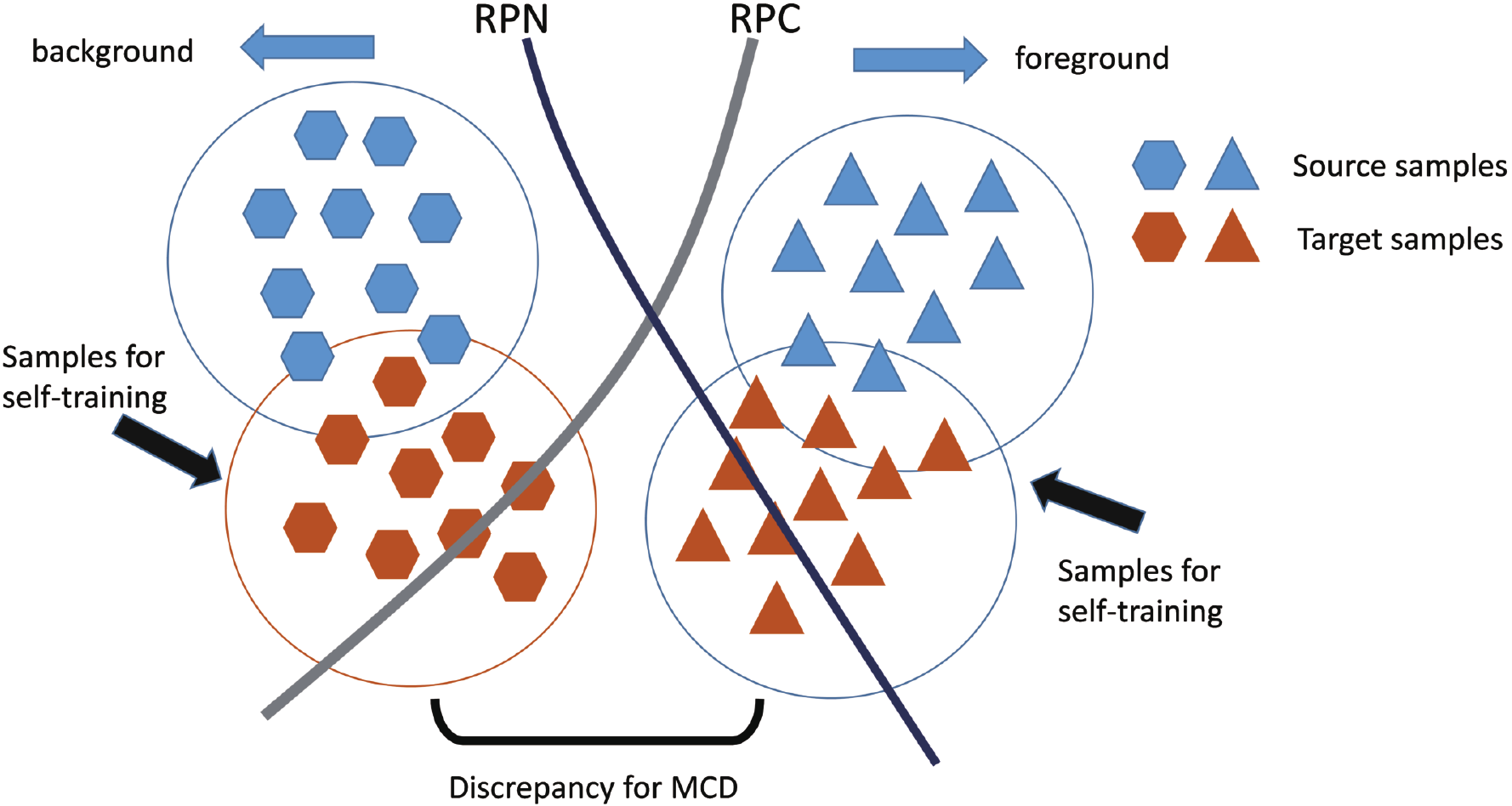}
		\end{center}
		\caption{Illustration of MCD principle. Samples far from the two decision boundaries of RPN and RPC tend to be reliable while others between the two boundaries are utilized for discrepancy computation.}
		\label{mcd}
	\end{figure}

	As described above, the loss term calculation of each ROI is multiplied by an adaptive weight during the collaborative self-training process. As shown in Fig.~\ref{weightcurve}, the weight value of RPN update depends on the relevant output score of the RPC branch, and vice versa. We design the weight function and guide the training process to focus more on ROIs with high-confidence. In this section, we creatively introduce a customized maximizing discrepancy classifier, i.e., MCD~\cite{saito2018maximum}, and point out that those ROIs with low-confidence can also be effectively leveraged to improve the model adaptation. MCD is a method originally proposed for domain adaptive image classification, which utilizes task-specific classifiers to align the distributions of source and target. It works by first separating the network into the feature extractor and classifier, and duplicating the latter. During training,  the two classifiers learn to maximize the prediction discrepancy between themselves while the feature extractor tries to minimize it. MCD theoretically pointed out that by minimizing and maximizing discrepancy, the transferability of the model can be effectively improved. We borrow it here and formulate the two-stage classification process in detection to satisfy its setting, which further complements the collaborative self-training. Specifically, we regard RPN and RPC as two foreground/background classifiers without duplication but weight ROIs in an opposite way. As the red curve shown in Fig.~\ref{weightcurve}, we assign higher weight to ROIs with low-confidence when performing MCD loss calculation. The discrepancy between RPN and RPC is defined as:
	
	\begin{equation}
	s_{cls}^{fg} = \sum_{c \in C}{s_{cls}^c},
	\end{equation}
	\begin{equation}
	L_{discrepancy}(s_{cls}, s_{rpn}) = | s_{cls}^{fg} - s_{rpn}^{fg} |,
	\end{equation} where \begin{math}C\end{math} is the set of foreground categories. \begin{math}
	L_{discrepancy}
	\end{math} measures the foreground/background discrepancy between RPN and RPC based on their predictions. In addition, we define the weight function as:
	\begin{equation}
	\begin{aligned}
	f_w(s_{cls}, s_{rpn}) =
	(2min(|s_{cls}^{fg}|, |1 - s_{cls}^{fg}|, |s_{rpn}^{fg}|, |1 - s_{rpn}^{fg}|))^\lambda.
	\end{aligned}
	\end{equation}
	\begin{math}f_w(s_{cls}, s_{rpn})\end{math} is set to obtain a larger value when both \begin{math}s_{cls}^{fg}\end{math} and \begin{math}s_{rpn}^{fg}\end{math} are around 0.5~(i.e., of low-confident prediction), and a smaller value when either of them approaches 0 or 1. It is introduced here to mitigate the negative impact of noisy RPN prediction in the calculation of MCD loss, which is defined as, 
	\begin{equation}
	L_{MCD} = f_w(s_{cls}, s_{rpn}) L_{discrepancy}(s_{cls}, s_{rpn}).
	\end{equation} The feature extractor \begin{math}F\end{math} is trained to minimize the \begin{math}
	L_{MCD}
	\end{math} while RPN and RPC try to maximize it alternately.
	As shown in Fig.~\ref{mcd}, each curve represents a decision boundary of a specific classifier. In this case, they are replaced by RPN and RPC. Samples between two decision boundaries are more likely to be wrongly classified while those far from decision boundaries are  more similar to the source domain and thus the output of which can be regarded as reliable pseudo labels.

	\subsection{RPN weighted alignment}
	Feature alignment between domains is a basic strategy often used by domain adaptive algorithms, and its effectiveness is widely recognized. As shown in Fig.~\ref{network}, we further introduce a domain discriminator to achieve cross-domain feature alignment, and verify that it stabilize and is complementary to the previously introduced collaborative self-training paradigm. However, due to the diversity of object category combinations and the scene complexity in object detection, simply aligning the whole image might lead to failure. 
	Some works have attempted to solve this problem, for example Saito~\etal\cite{saito2018strong} use a weak global alignment and Zhu~\etal \cite{zhu2019adapting} try to align the ROIs after region mining. We instead design a fine-grained and more flexible alignment with RPN score, i.e., the discriminator focusing on the regions of higher foreground probability. 
	
	Specifically, we design RPN weighted alignment as a local domain discriminator\cite{saito2018strong}. Discriminator \begin{math}D\end{math} takes the feature of size \begin{math}H \times W \times C\end{math} from the backbone as input, and outputs a \begin{math}H \times W\end{math} probability map, the value of which denotes the domain probability of the specific position. We scale the RPN foreground map  \begin{math}f\end{math} to the same spatial size (\begin{math}H \times W \times 9\end{math}) and weight the loss as follows:
	
	\begin{equation}
	\label{weight_local_1}
	L_{adv_s} = \frac{1}{HW}\sum_{w=1}^{W}\sum_{h=1}^{H}
	(D(F(x_s))_{wh} ^2 \sum_{i=1} ^ {9}(f_{i})_{wh}),
	\end{equation}
	
	\begin{equation}
	\label{weight_local_2}
	L_{adv_t} = \frac{1}{HW}\sum_{w=1}^{W}\sum_{h=1}^{H}
	((1 - D(F(x_t))_{wh}) ^2 \sum_{i=1} ^ {9}(f_{i})_{wh}).
	\end{equation}
	
	\begin{math}f_i\end{math} is the i-th channel of the RPN foreground map~\begin{math}f\end{math} with size~\begin{math}H \times W\end{math}, which represents the probability that the i-th anchor box defined at each position belongs to the foreground. \begin{math}(f_i)_{wh}\end{math} and \begin{math}D(F(x))_{wh}\end{math} are the element of \begin{math}f_i\end{math} and~\begin{math}D(F(x))\end{math} at position \begin{math}(w,h)\end{math}. The foreground map \begin{math}f\end{math} might be rough at the beginning, but it will continue to be optimized as the collaborative self-training iterates and become an effective complement to the collaborative self-training at the backbone. 

	\subsection{Overall objective}
	\label{overall}
	The detection loss of original Faster RCNN consists of localization loss \begin{math}L_{rpn}\end{math} calculated on RPN and classification loss \begin{math}L_{cls}\end{math} on RPC. For source image, loss is defined as follows:
	
	\begin{equation}
	L_s = L_{rpn} + L_{cls} + L_{adv_s}.
	\end{equation}
	
	For the target domain, it is slightly different due to $L_{MCD}$. we define the backbone loss and the loss of RPN and RPC as:
	
	\begin{equation}
	L_{t_{backbone}} = L_{adv_t} + \alpha L_{rpn_t} + \beta L_{cls_t} + \gamma L_{MCD},
	\end{equation}
	
	\begin{equation}
	L_{t_{RPN, RPC}} = \alpha L_{rpn_t} + \beta L_{cls_t} - \gamma L_{MCD},
	\end{equation}
	and the loss for the discriminator is:
	
	\begin{equation}
	L_{D} = L_{adv_s} + L_{adv_t}.
	\end{equation}
	
	\begin{math}\alpha, \beta, \gamma\end{math} control the trade-off between the detection loss of the source image and other losses. As~\cite{ganin2016domain}, we adopt GRL (gradient reverse layer), which flips the sign of gradients in back-propagation, to implement the adversarial loss. 

	\section{Experiment}
	\label{experimentsection}
	We adopt unsupervised domain adaptation protocol and use four datasets in our experiments for evaluation, including Cityscapes~\cite{cordts2016cityscapes}, FoggyCityscapes~\cite{sakaridis2018semantic}, Sim10k~\cite{johnson2016driving} and KITTI~\cite{geiger2012we}. Both images and annotations are provided for the source domain, while only images are available for target domain at training. As with~\cite{chen2018domain} and~\cite{zhu2019adapting}, we evaluate our method on three kinds of domain shifts.
	
	\subsection{Implement details}
	\label{implement}
	
	Following~\cite{chen2018domain},~\cite{saito2018strong} and~\cite{zhu2019adapting}, we adopt Faster RCNN~\cite{ren2015faster} with VGG16~\cite{simonyan2014very} backbone and ROI-alignment~\cite{he2017mask} in all our experiments. We resize the input images so that the length of the shorter size is 600 and keep the aspect ratio following~\cite{saito2018strong}. Our model is trained with three steps using SGD with 0.9 momentum and 0.001 learning rate. We first pre-train the model using the source dataset, followed by 10,000 iterations with \begin{math}L_{det}\end{math} calculated on the source domain and \begin{math}L_{adv}\end{math} on both domains, and finally train the network with all loss terms in section~\ref{overall} for 6,000 iterations. Without specific notation, we set \begin{math}\alpha\end{math} as 0.1, \begin{math}\beta\end{math} as 0.05, \begin{math}\gamma\end{math} as 0.1. 
	For simplicity, pseudo boxes with confidence under 0.9 are discarded in RPN self-training and \begin{math}f_w(s_{cls})\end{math} is set to 1. \begin{math}\lambda\end{math} is set to 2 in \begin{math}L_{MCD}\end{math} and 5 in other loss terms. 
	We implement all methods with Pytorch~\cite{paszke2017automatic}. The architecture of domain discriminator follows~\cite{saito2018strong}.

	We compare our method with four baselines: Faster RCNN~\cite{ren2015faster}, domain adaptive Faster RCNN (DA-Faster)~\cite{chen2018domain}, Strong Weak alignment (SWDA)~\cite{saito2018strong}, and selective cross-domain alignment (SCDA)~\cite{zhu2019adapting}. The Faster RCNN model is only trained with the source images and labels without referring to the target data, which is also referred to as the source only model.

	
	
	

	\subsection{Domain Adaptation for Detection}
	\paragraph{\textbf{Normal to Foggy}}
	Cityscapes~\cite{cordts2016cityscapes} is used as the source domain while FoggyCityscapes\cite{sakaridis2018semantic} as the target. In both domains, we use the training set for pre-training and adaptation without augmentation, and evaluate our model on the validation set for 8 classes. The results are reported in Table~\ref{cityscapestofoggy}. As shown in the table, our method outperforms the existing state-of-the-art by 1.6\% w.r.t mAP. Besides, our method outperforms existing methods in class \begin{math}car\end{math}, which is the most common object in target domain. 
	
	\paragraph{\textbf{Synthetic to Real}}
	
	Sim10k\cite{johnson2016driving} is used as the source domain and Cityscapes as the target domain. Similar to~\cite{chen2018domain},~\cite{saito2018strong} and~\cite{zhu2019adapting}, we evaluate the detection performance on~\begin{math}car\end{math} in Cityscapes validation set. The results of our method is reported in Table~\ref{sim10ktocityscapes}. Our method outperforms the existing state-of-the-art method by 1.5\% w.r.t mAP.
	
	\paragraph{\textbf{Cross Camera Adaptation}}
	We used KITTI\cite{geiger2012we} as source domain and Cityscapes as the target domain for evaluation. 
	The results under this scenario is reported in Table \ref{kittitocityscapes}. Our method improves the existing best method by 1.1\% w.r.t mAP.

	\begin{table}
		\caption{AP(\%) from Cityscapes to FoggyCityscapes}
		\begin{center}
			\label{cityscapestofoggy}
			\begin{tabular}{lccccccccc}
				\hline
				Method & person & rider & car & truck & bus & train & motobike & bicycle & mAP \\
				\hline
				Faster RCNN \cite{ren2015faster} & 29.7 & 32.2 & 44.6 & 16.2 & 27.0 & 9.1 & 20.7 & 29.7 & 26.2 \\
				
				DA-Faster \cite{chen2018domain} & 25.0 & 31.0 & 40.5 & 22.1 & 35.3 & 20.2 & 20.0 & 27.1 & 27.6 \\
				
				SWDA \cite{saito2018strong} & 29.9 & 42.3 & 43.5 & 24.5 & 36.2 & \textbf{32.6} & 30.0 & 35.3 & 34.3 \\
				
				SCDA \cite{zhu2019adapting} & $\textbf{33.5}$ & 38 & 48.5 & $\textbf{26.5}$ & 39 & 23.3 & 28 & 33.6 & 33.8 \\
				
				Proposed & 32.7 & $\textbf{44.4}$ & $\textbf{50.1}$ & 21.7 & \textbf{45.6} & 25.4 & \textbf{30.1} & \textbf{36.8} & \textbf{35.9}\\
				\hline
			\end{tabular}
		\end{center}
	\end{table}

	\begin{table}[!htb]
		\centering
		\begin{minipage}[t]{0.45\textwidth}
			\centering
			\makeatletter\def\@captype{table}\makeatother\caption{AP on ``Car''(\%) from Sim10k to Cityscapes.}
			\label{sim10ktocityscapes}
			\begin{tabular}{lc}
				\hline
				Method & AP on ``Car'' \\
				\hline
				Faster RCNN \cite{ren2015faster} & 34.57 \\
				DA-Faster \cite{chen2018domain} & 38.97 \\
				SWDA \cite{saito2018strong} & 40.10 \\
				SCDA \cite{zhu2019adapting} & 43.05\\
				
				Proposed & 44.51 \\
				\hline
			\end{tabular}
		\end{minipage}
		\begin{minipage}[t]{0.45\textwidth}
			\centering
			\makeatletter\def\@captype{table}\makeatother\caption{AP on ``Car''(\%) from KITTI to Cityscapes.}
			\label{kittitocityscapes}
			\begin{tabular}{lc}
				\hline
				Method & AP on ``Car'' \\
				\hline
				Faster RCNN\cite{ren2015faster} & 34.9 \\
				DA-Faster \cite{chen2018domain} & 38.5 \\
				SWDA \cite{saito2018strong} & - \\
				SCDA \cite{zhu2019adapting} & 42.5\\
				Proposed & 43.6 \\
				\hline
			\end{tabular}
		\end{minipage}
	\end{table}

	\section{Ablation Study}
	\label{RPN_training_ablation}
\begin{figure}
		\begin{subfigure}[t]{.24\textwidth}
			\centering
			\includegraphics[width=\linewidth]{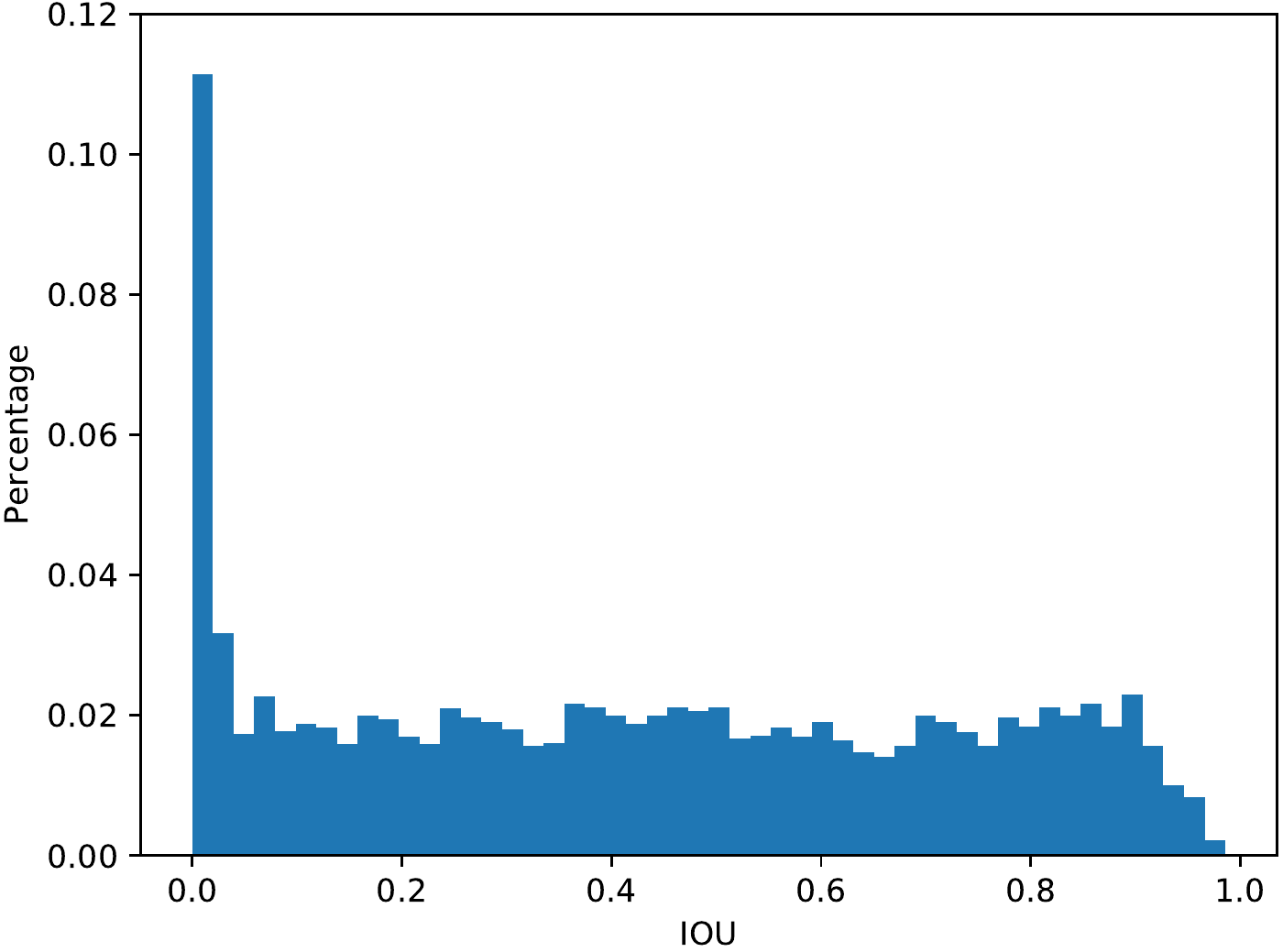}
			\caption{Source Only}
		\end{subfigure}
		\begin{subfigure}[t]{.24\textwidth}
			\centering
			\includegraphics[width=\linewidth]{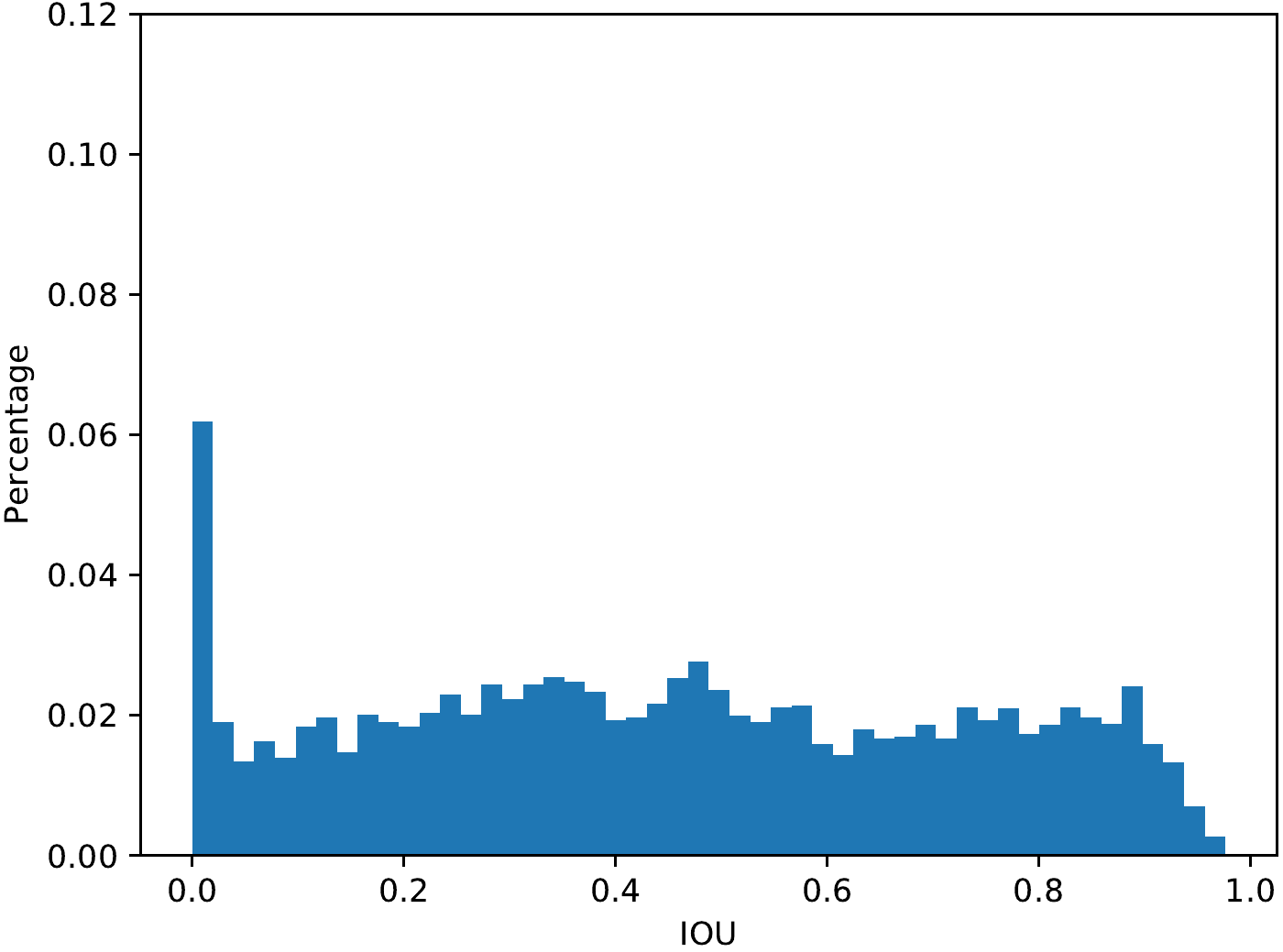}
			\caption{Naive alignment}
		\end{subfigure}
		\begin{subfigure}[t]{.24\textwidth}
			\centering
			\includegraphics[width=\linewidth]{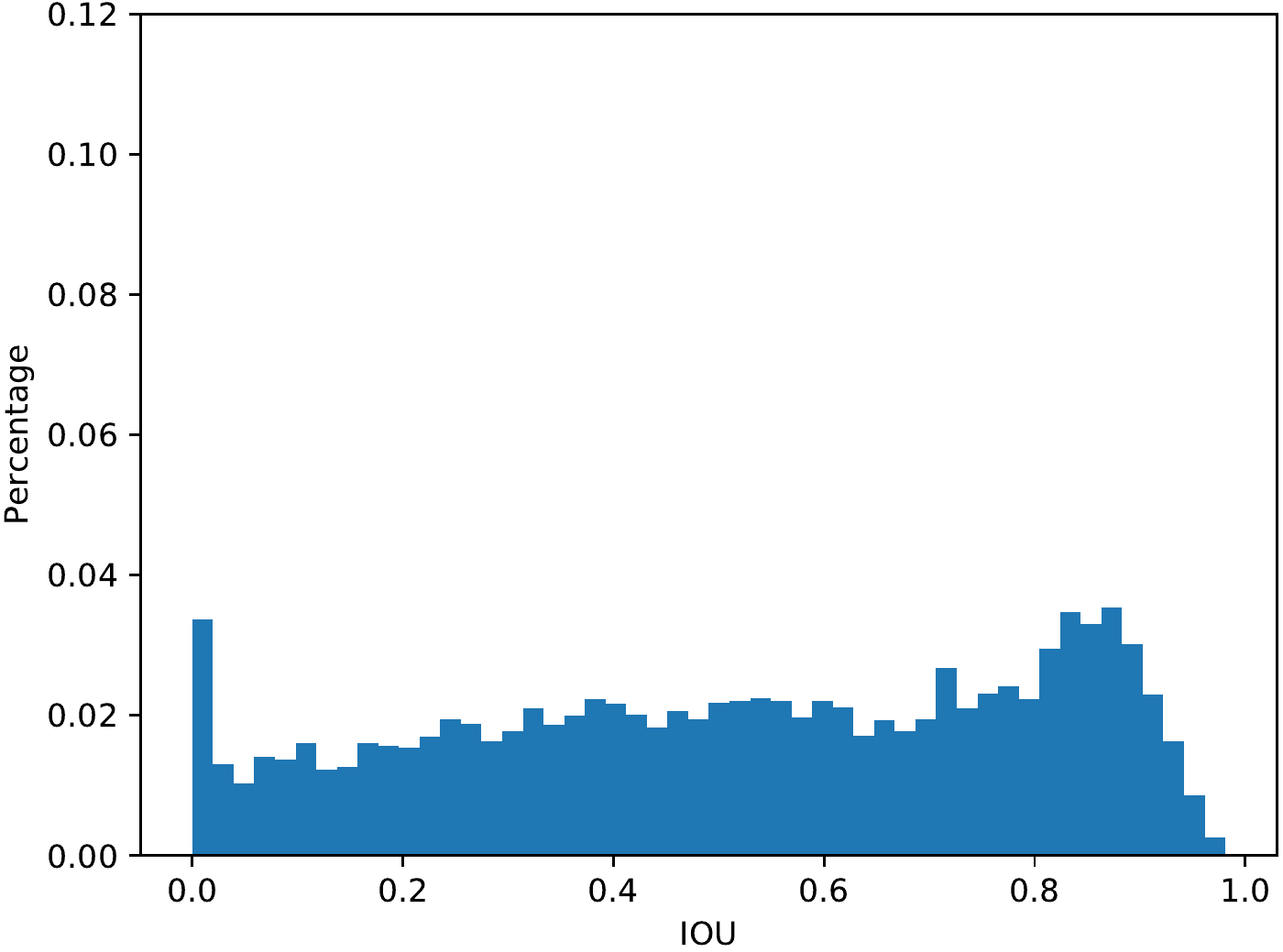}
			\caption{RPN adaption}
		\end{subfigure}
		\begin{subfigure}[t]{.24\textwidth}
			\centering
			\includegraphics[width=\linewidth]{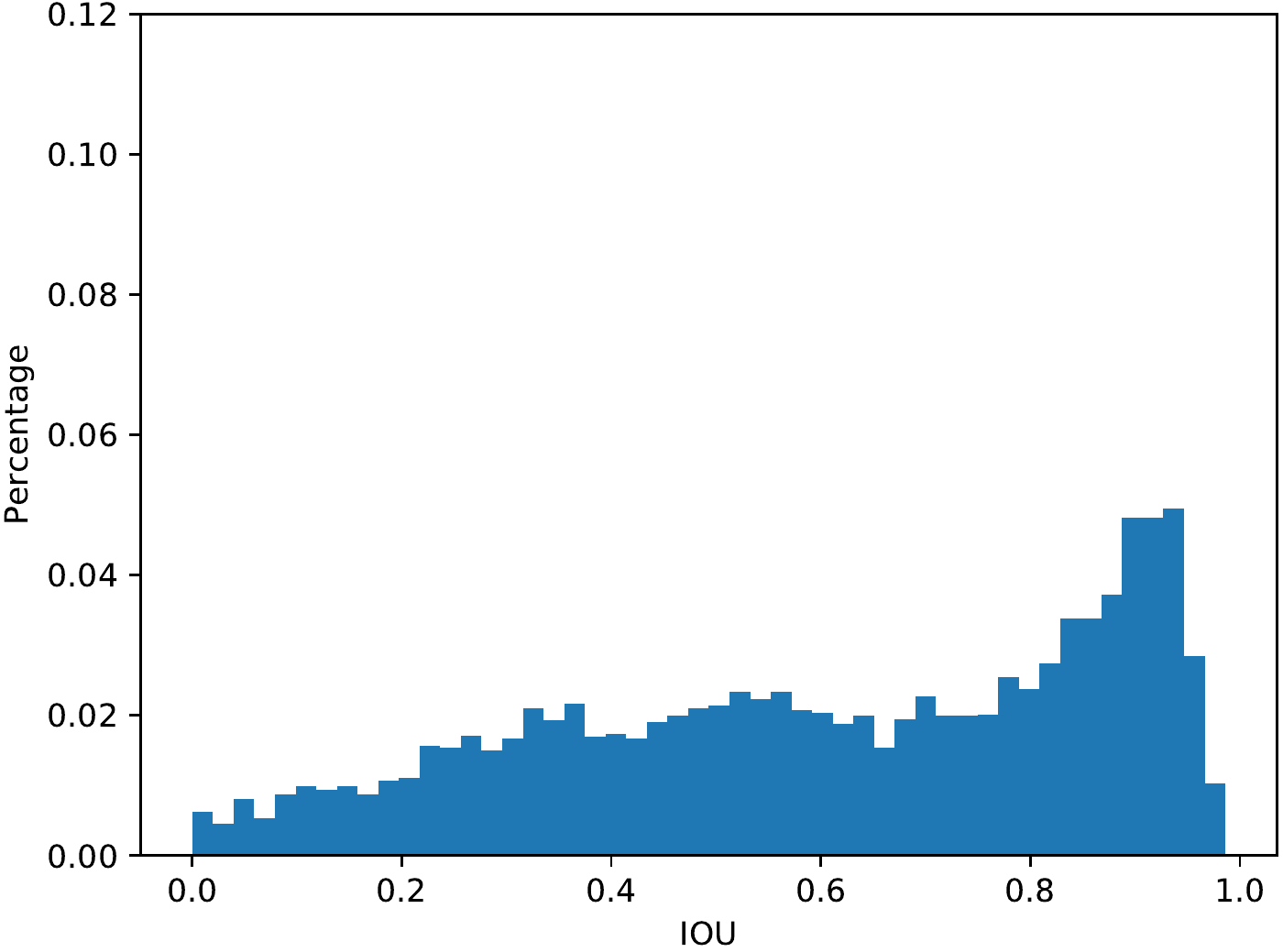}
			\caption{Oracle}
		\end{subfigure}
		\caption{Proposal coverage distribution of different proposal generation models. ``Naive alignment'' stands for non-weighted local alignment and ``Oracle'' refers to the model trained with labeled target dataset. RPN adaption is trained using our proposed collaborative self-training paradigm. 
		}
		\label{fig:pcdistribution}
	\end{figure}

	\paragraph{\textbf{Effectiveness of RPN adaptation}} 
	As one of our core motivations is to explore the significance of the RPN module for domain adaptation. We verify the superiority of collaborative self-training by analyzing the quality of region proposals generated by different adaptation models. We first define a metric called~\begin{math}proposal\end{math} \begin{math}coverage\end{math}. Given a ground truth bounding box, we define the largest IOU with all detected proposals as its proposal coverage. For each ground truth  in the target domain, we calculate the proposal coverage and count the distribution for each detection model. 
	We conduct experiments on the adaptation from Sim10k to the Cityscapes dataset for comparison. Firstly, in order to verify that the naive local alignment in the backbone branch is not sufficient for domain adaptive object detection, we adopt a non-weighted local alignment method as naive alignment for comparison, in which a domain discriminator is applied to every position of the feature map. It can be implemented by removing the \begin{math}f_i\end{math} weighting strategy in Eq~\ref{weight_local_1} and Eq~\ref{weight_local_2}. We compare the proposal coverage distribution of four different detection models, including the source-only model, naive alignment model, our RPN adaptation with collaborative self-training and the Oracle model~(i.e., the performance upper bound which refers to the annotations of the target domain). As shown in Fig.~\ref{fig:pcdistribution}, our method greatly improves the quality of the generated object proposals in the target domain and its proposal coverage distribution is much more closer to the Oracle model, when compared with the naive alignment. Specifically, our method greatly reduces the boxes with proposal coverage = 0 compared with both the source only and naive alignment models. It also obviously improves the quality of proposals with IOU$\ge$0.5 while the naive alignment mainly changes the distribution of IOU$<$0.5 w.r.t the source-only baseline. This shows that our method can effectively improve the accuracy of the generated proposals, and therefore bring about significant numerical performance improvements~(Table~\ref{ablationtable}). Noted that the performance benefit may become more apparent as the IOU threshold setting to a higher value.
	
	\paragraph{\textbf{Effectiveness of Different Components}}
	
	We evaluate the contribution of different components by designing several variants of our model. The results are reported in Table~\ref{ablationtable}. All experiments are conducted on the adaptation from Sim10k to Cityscapes dataset. Hyper-parameter setting of all model variants remain the same as described in Section~\ref{implement}. As shown in the table, incorporating the core module, i.e., collaborative self-training~(CST), to the baseline source only model can bring significant performance gain of 7.76\% w.r.t AP, increasing AP from 34.57\% to 42.33\%. This reveals that the domain adaptation capability can indeed be improved by effectively mining the complementary advantages of RPN and RPC. On the other hand, applying a naive local alignment only results in 2.46\% performance improvement, which proves that simple alignment on the backbone branch is far from sufficient for domain adaptive object detection. Nevertheless, the proposed weighted local alignment still outperforms the naive alignment by 1.28\% w.r.t AP even without RPN self-training. It is worth noting that using MCD alone does not significantly improve the baseline~(36.42\% VS 34.57\%) because most of the uncertain ROIs are filtered out by RPN. Last but not least, as can be seen from the last two rows of the table, CST is complementary to the RPN weighted alignment and MCD, our entire model gains an additional 2.18\% AP when compared to the CST-only version. In general, all three proposed components make their own contributions compared with the source-only model, which overall improve the baseline by 9.94\% w.r.t AP.

	\begin{figure*}[h]
		\begin{center}
			\begin{subfigure}[t]{.31\textwidth}
				\centering
				\includegraphics[width=\linewidth]{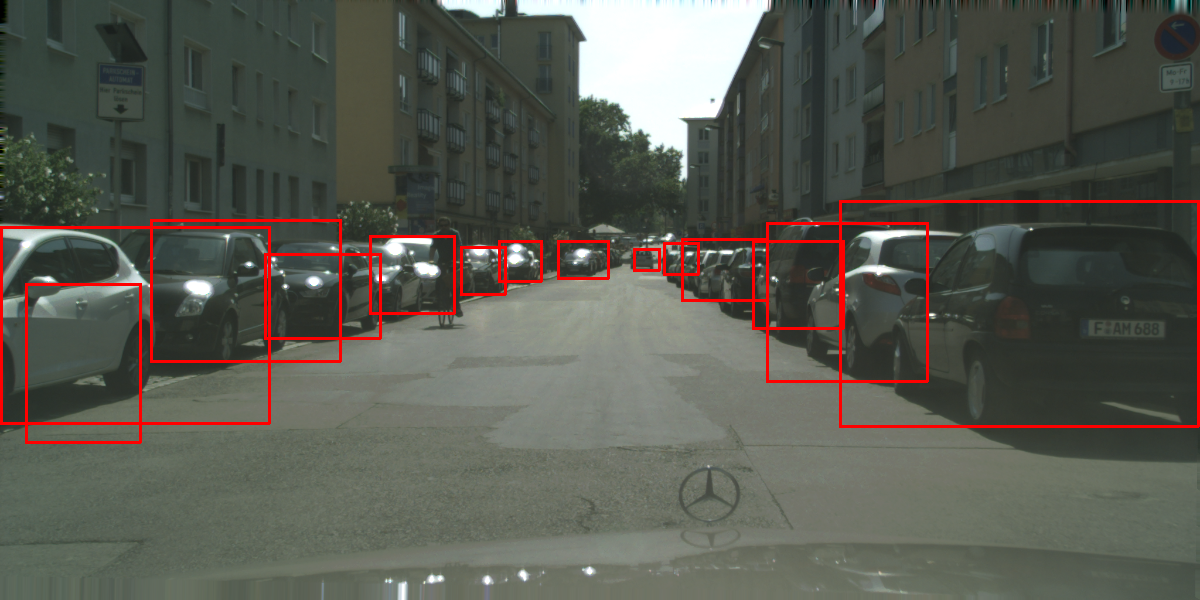}
			\end{subfigure}
			\begin{subfigure}[t]{.31\textwidth}
				\centering
				\includegraphics[width=\linewidth]{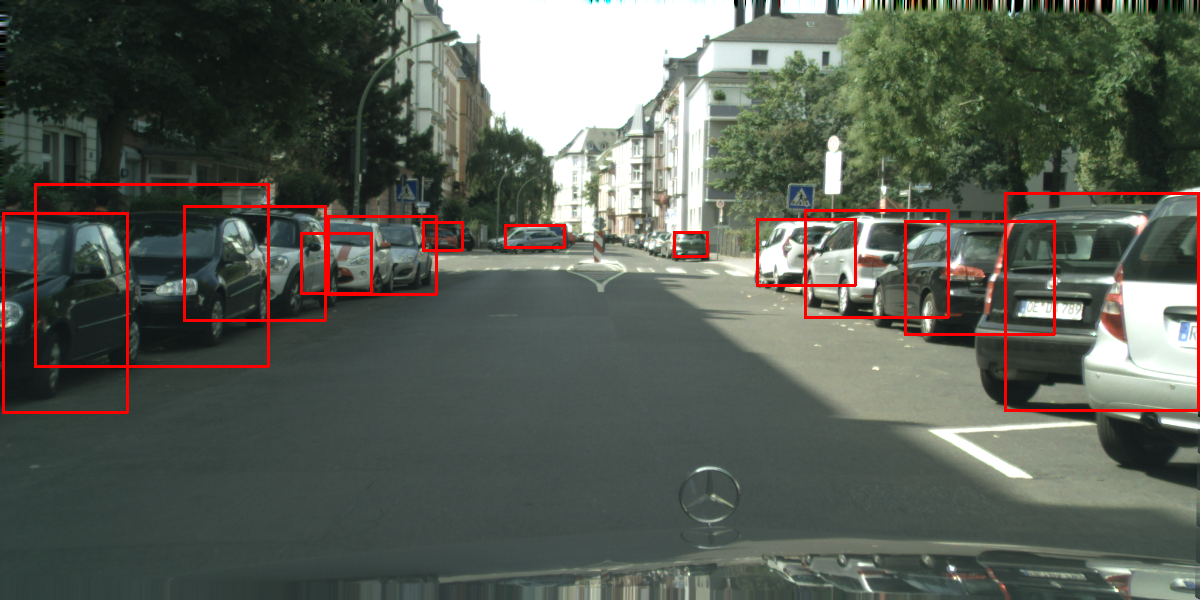}
			\end{subfigure}
			\begin{subfigure}[t]{.31\textwidth}
				\includegraphics[width=\textwidth]{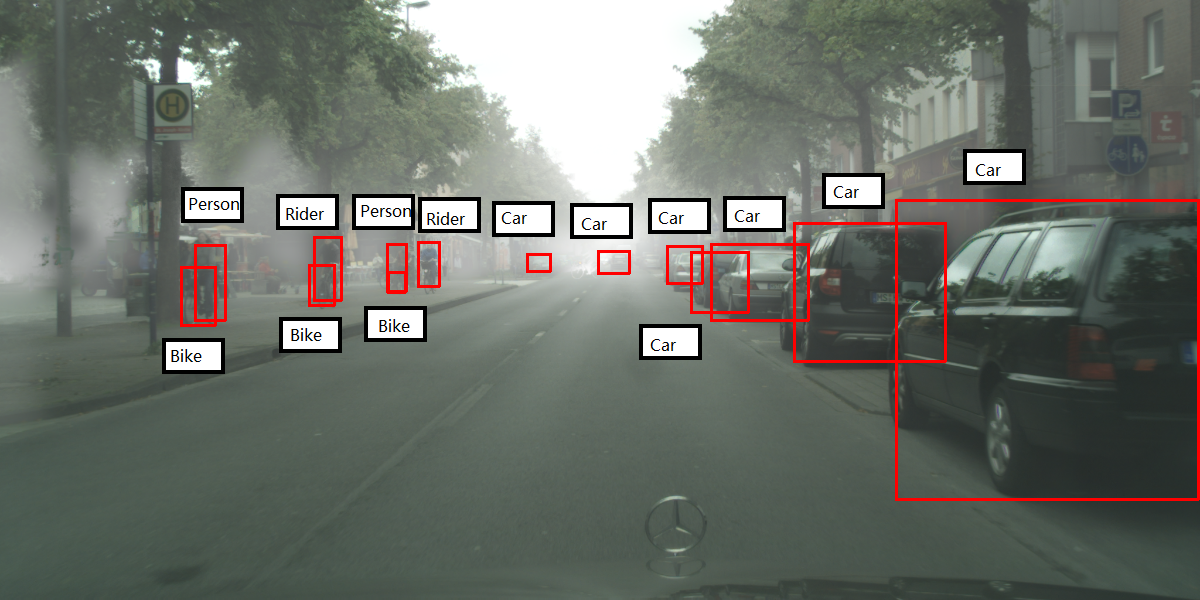};
			\end{subfigure}

			\begin{subfigure}[t]{.31\textwidth}
				\centering
				\includegraphics[width=\linewidth]{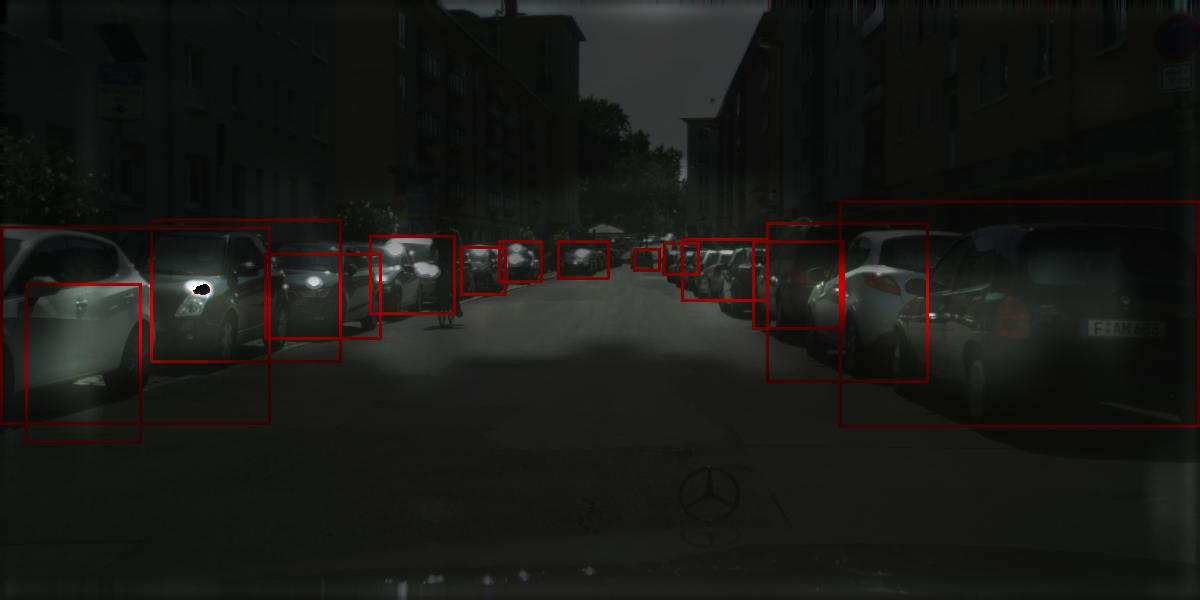}
			\end{subfigure}
			\begin{subfigure}[t]{.31\textwidth}
				\centering
				\includegraphics[width=\linewidth]{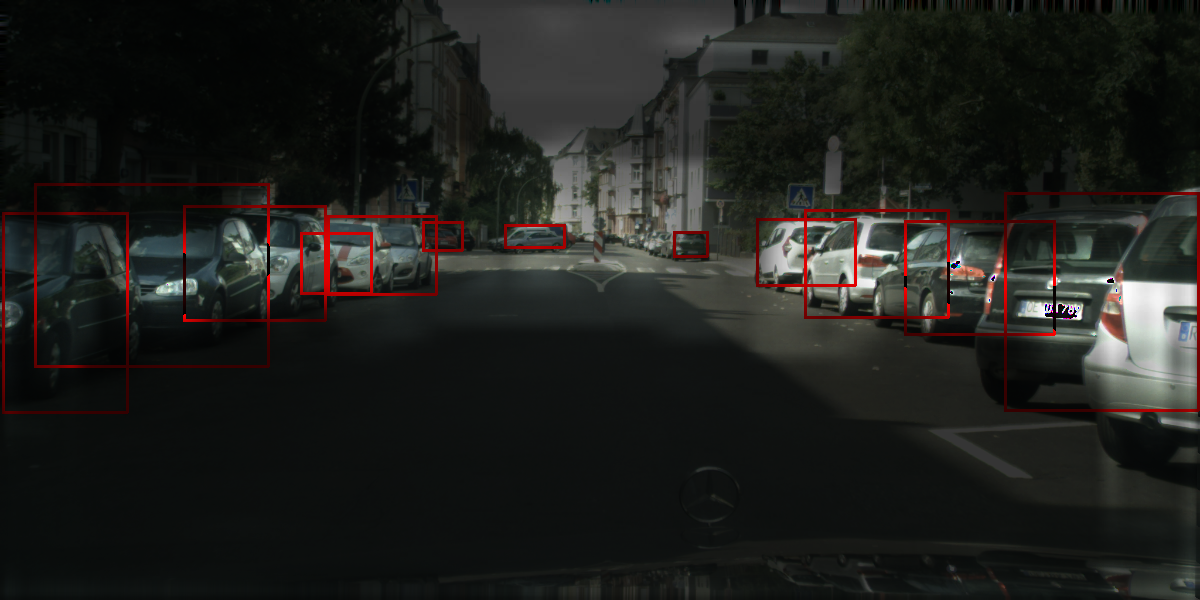}
			\end{subfigure}
			\begin{subfigure}[t]{.31\textwidth}
				\centering
				\includegraphics[width=\linewidth]{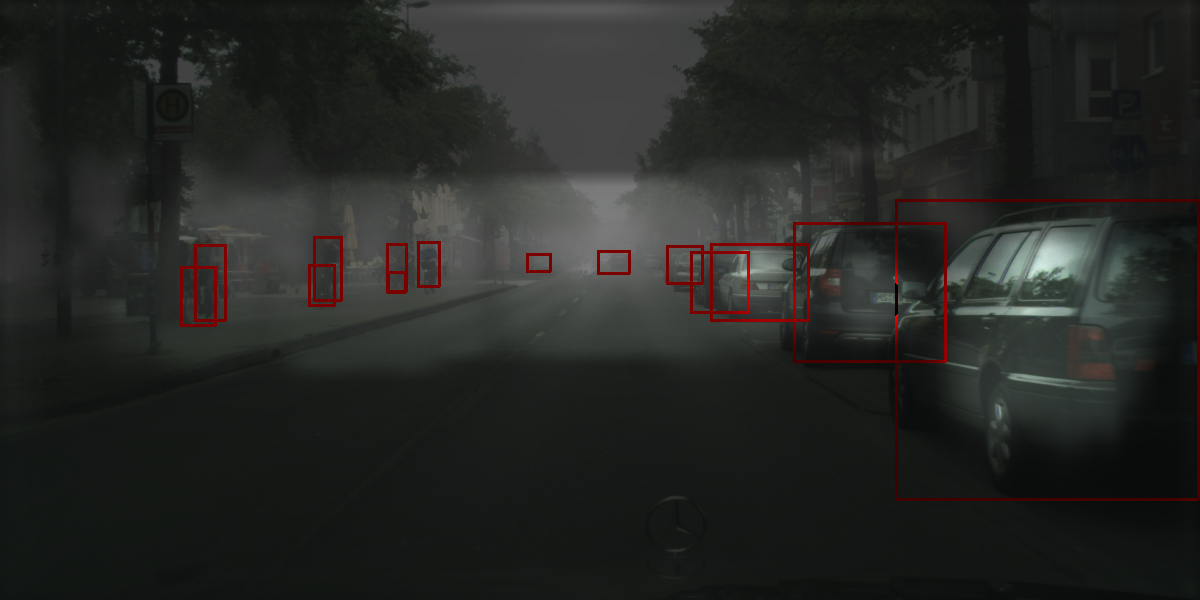}
			\end{subfigure}
			\caption{Detection result on target domain. From left to right: Sim10k to Cityscapes; Kitti to Cityscapes; Cityscapes to FoggyCityscapes. The second row shows the weight inferred from our RPN weighted local domain discriminator. Brighter colors indicate higher attention. Apparently,  regions with considered objects (e.g. car) are of higher weight in loss calculation.}
			\label{fig_resultshow}
		\end{center}
	\end{figure*}
	
	\begin{table}
		\caption{AP on ``Car'' from Sim10k to Cityscapes of different method variants to demonstrate the effectiveness of the proposed algorithm. ``Local'' represents naive local alignment and ``CST'' refers to the collaborative self-training.}
		\begin{center}
			\label{ablationtable}
			\begin{tabular}{ccccc}
				\hline
				Local & Weight Local & CST & MCD & AP on Car \\
				\hline\hline
				& & & & 34.57 \\
				\hline
				\checkmark & & & & 37.03 \\
				\hline
				& \checkmark & & & 38.31 \\
				\hline
				& & \checkmark & & 42.33 \\
				\hline
				& & & \checkmark & 36.42 \\
				\hline
				& \checkmark & \checkmark && 43.08 \\
				\hline
				& \checkmark & \checkmark & \checkmark & 44.51 \\
				\hline
			\end{tabular}
		\end{center}
	\end{table}

	\paragraph{\textbf{Visualization of detection results}}	
	The detection results after adaptation are illustrated in Fig.~\ref{fig_resultshow}. Our model can accurately localize and classify objects under different kinds of domain shifts. We also visualize the weight used in the local alignment calculation. It is obvious that RPN weighted method can effectively suppress non-critical parts of the image. As shown in the Figure, although the sky and roads occupy most of the area of the image, the inferred weight map shows that these areas have little effect on distinguishing objects of different domains, which is consistent with our optimization goal.


	\section{Conclusion}
	In this paper, 
	we are the first to empirically reveal that the RPN and RPC module in the endemic two-stage detectors (e.g., Faster RCNN) demonstrate significantly different transferability when facing large domain gap. Base on this observation, we design a collaborative self-training method for RPN and RPC to train each other with ROIs of high-confidence. Moreover, a customized maximizing discrepancy classifier is introduced to effectively leverage ROIs with low-confidence to further increase the accuracy and generalization of the detection model. 
	Experimental results demonstrated that our method significantly improves the transferability and outperforms existing methods in various domain adaptation scenarios.
	
	\section*{Acknowledgements}
	This work was supported in part by the Guangdong Basic and Applied Basic Research Foundation (2020B1515020048), in part by the National Natural Science Foundation of China (61976250, 61702565, U1811463), in part by the National High Level Talents Special Support Plan (Ten Thousand Talents Program), in part by the Fundamental Research Funds for the Central Universities (18lgpy63). This work was also sponsored by Meituan-Dianping Group.

	\clearpage
	%
	%
	\bibliographystyle{splncs04}
	\bibliography{egbib}
\end{document}